\documentclass[11pt]{article}

\usepackage[final]{acl}

\usepackage{times}
\usepackage{latexsym}

\usepackage[T1]{fontenc}

\usepackage[utf8]{inputenc}

\usepackage{microtype}

\usepackage{inconsolata}

\usepackage{graphicx}

\usepackage{amsmath,amssymb}

\usepackage{hyperref}

\usepackage{booktabs}
\usepackage{multirow}
\usepackage{array}
\usepackage{caption}
\usepackage{subcaption}
\usepackage{float}

\newcommand{\softmax}{\mathrm{softmax}}
\newcommand{\cosim}{\mathrm{cos}}
\newcommand{\KL}{\mathrm{KL}}

\title{Beyond Neural Incompatibility: Cross-Scale Knowledge Transfer in Language Models through Latent Semantic Alignment}

\author{Jian Gu \\
  Monash University \\
  \texttt{jian.gu@monash.edu} \\\And
  Aldeida Aleti \\
  Monash University \\
  \texttt{aldeida.aleti@monash.edu} \\\AND
  Chunyang Chen \\
  Technical University of Munich \\
  \texttt{chun-yang.chen@tum.de} \\\And
  Hongyu Zhang \\
  Chongqing University \\
  \texttt{hyzhang@cqu.edu.cn} \\
}

\usepackage{mystyle}

\begin{document}
\maketitle

\begin{abstract}
Language Models (LMs) encode substantial knowledge in their parameters, yet it remains unclear how to transfer such knowledge in a fine-grained manner, namely parametric knowledge transfer (PKT). A central challenge is to make cross-scale transfer effective and efficient when source and target models differ in architecture and parameterization, making direct parameter reuse strongly limited by neural incompatibility.
In this paper, we identify latent semantic alignment as the key prerequisite for cross-scale knowledge transfer. Instead of directly moving layer parameters, our approach uses activations as the transfer medium. \textsc{SemAlign} has two stages: an \emph{layer attribution} stage that attributes task-relevant source layers and selects exactly one source layer for each target layer, and a \emph{semantic alignment} stage that pairs them layer by layer and optimizes the target with source-side semantic supervision. The alignment is carried out in latent space through semantic decomposition and recomposition. During the shallow-to-deep transfer, only the frontier target layer is trainable. The layer objective supervises the residual contribution of that layer by matching centered token-token relation geometry against an aligned supervisory residual, while output KL preserves source-level predictive behavior. The transferred medium is therefore neither a parameter block nor an absolute hidden state, but target-space residual geometry induced by paired source-layer supervision.
Evaluations on four benchmarks demonstrate the efficacy of \textsc{SemAlign}, and further analysis confirms that semantic decomposition and recomposition provide a stable mechanism for cross-scale knowledge transfer.
\end{abstract}


\section{Introduction}
\label{sec:introduction}

Language Models (LMs) store substantial knowledge in their parameters, and recent work has shown that such knowledge can be located, traced, edited, or transferred. A natural objective is therefore parametric knowledge transfer (PKT). The goal is to move internal knowledge from a stronger and larger source to a weaker and smaller target so that the target acts more like the source. This objective is especially important for cross-scale transfer, where the source and target differ in depth, width, and parameterization. However, direct parameter reuse or injection is strongly constrained by neural incompatibility: components that are useful in the source parameter space need not have a compatible meaning in the target parameter space \citep{tan2025neural}.

This motivates a transfer setting in which a target receives source-side internal signals directly, with less loss and lower dependence on textual approximation. We argue that cross-scale PKT should be semantics-first. Rather than treating source weights as the transferable medium, we treat activations as the medium of transfer and align semantics before any parameter updates. This view is motivated by the observation that layer activations expose task-specific computation in a form that is closer to model behavior than raw parameters, while still being compact enough for efficient transfer.

We propose \textsc{SemAlign}, a semantics-based PKT method that uses layer outputs as the transfer signal. Compared with common extract-align-inject PKT pipelines, SemAlign introduces a two-stage workflow:
First, we run layer attribution only on the source to locate layers that carry task-relevant signal, and we keep exactly one selected source layer for each target layer;
Second, we perform a semantic-alignment stage: the selected source layers are paired with target layers from shallow to deep, their hidden states are decomposed into semantic components in the source space and recomposed as target-space supervisory states, and the target is optimized at those paired interfaces.
Semantic alignment follows a shallow-to-deep frontier schedule. At each frontier stage, only the current target layer is trainable and all other target layers are frozen. The paired-layer objective supervises the target layer's residual contribution by matching centered token-token relation geometry against an aligned supervisory residual, while an output KL term preserves source-level predictive behavior. Unlike conventional intermediate-layer distillation, which directly matches source features in the source space, SemAlign first transports the supervisory signal into the target semantic space and then optimizes the target residual relation structure. By aligning how model layers behave, SemAlign reduces neural incompatibility while keeping the procedure simple and efficient.

We evaluate SemAlign under the same setup as prior work \citep{tan2025neural}. We use four standard benchmarks on professional knowledge, mathematical reasoning, and code generation. The experiments are conducted with Llama2 models \citep{touvron2023llama}, performing task-related PKT by pairing larger sources with smaller targets that differ in depth and width. Across all tasks, SemAlign improves target performance over task-matched baselines and over parameter-space transfer baselines.
We have two main findings:
First, our attribution-then-alignment approach yields stable cross-scale knowledge transfer without requiring an additional injection stage;
Second, shallow-to-deep frontier optimization is sufficient to induce broader behavioral alignment, which keeps the method efficient in both compute and data.
The replication repository is attached as supplementary material.

To summarize, our contributions are as follows:
\begin{itemize}
\item We present a semantics-first view of parametric knowledge transfer for cross-scale LMs. The formulation treats the latent semantics between paired layers as the medium for knowledge transfer, instead of model parameters.
\item We introduce \textsc{SemAlign}, which combines layer attribution and pairing, latent semantic alignment, and residual relation loss for knowledge transfer. It mitigates limitations associated with neural incompatibility.
\item We conducted experiments on representative benchmarks with Llama2-family models. Results and analysis identify the key factors that ease cross-scale knowledge transfer and show consistent gains in transfer efficacy.
\end{itemize}

\section{Related Work}
\label{sec:related_work}

\subsection{Knowledge Attribution in Language Models}
Knowledge attribution studies methods for identifying where knowledge resides in large language models and how those components influence predictions.
The focus has moved from layer-level inspections to neuron-level and path-level analyses that scale to large models.
One representative line designs a static, single-pass neuron score that separates ``query'' and ``value'' neurons and avoids repeated gradient passes~\citep{yu2024neuron}.
Moving from units to mechanisms, information-flow routes rebuild prediction-time computation as a sparse graph and show how influential parts work together during inference~\citep{ferrando2024information}.
In practical analyses, \textsc{Captum} provides operators for layer and neuron attribution, including Internal Influence, Neuron Integrated Gradients, and DeepLIFT or SHAP, which many studies adopt as reproducible baselines~\citep{kokhlikyan2020captumunifiedgenericmodel}.
Recent evidence also reports degenerate knowledge neurons, where different neuron sets encode the same fact; this observation supports concept-aware or path-aware selection when using attribution to guide editing or transfer~\citep{chen2025cracking}.

\subsection{Semantic Analysis and Latent Space Alignment}
Semantic analysis and latent-space alignment shape and match internal representations so that model adaptation preserves meaning rather than only optimizing an output loss.
Within this view, two connected methods motivate our use of semantic bases.
Vocabulary Defined Semantics (VDS) uses the model vocabulary to anchor directions in the hidden space and then clusters examples around these anchors, which stabilizes in-context learning by better matching data to the model's internal semantic frame~\citep{gu2024vocabulary}.
Building on that foundation, Semantic Aware Layer Freezing (SALF) treats the structure exposed by VDS as semantic anchors at the layer level and freezes those parts while tuning the remainder, which preserves core semantics and works with parameter-efficient finetuning and quantization~\citep{gu2025semantic}.
A complementary research thread adjusts hidden states at test time with small edits, showing that behavior can be steered through representation space without heavy retraining~\citep{kong2024aligning}.

\subsection{Parametric Knowledge Transfer}
Knowledge transfer includes source-target distillation, representational matching across layers, and parameter mixing through model merging or task vectors.
These approaches provide strong baselines and tools, yet they often work in the output space or assume closely related architectures~\citep{xu2024survey,yang2025survey,yang2026model}. Classical distillation and hint-based supervision can also use intermediate features, but they typically match source distributions or source features directly in the source space. Our setting instead constructs a target-space supervisory signal from semantic bases, so the optimization remains tied to representation transfer across potentially incompatible latent spaces.
Recent studies frame the problem as parametric knowledge transfer, where the goal is to move internal knowledge that lives inside a model, including parameters and intermediate computations such as activations and residual streams.
A representative system, \textsc{Seeking}, extracts sensitive components from a source, injects them into a target through LoRA initialization, and then applies post-alignment finetuning; results indicate that cross-scale transfer is feasible and that alignment quality is important for stability~\citep{zhong2024seeking}.
Follow-up work \textsc{LaTen} on Neural Incompatibility examines alignment as the main bottleneck across scales and distinguishes two design choices: PostPKT, which follows extract, inject, and train, and PrePKT, exemplified by LaTen, which aligns parametric spaces with light training before transfer~\citep{tan2025neural}.
Our method leverages LM semantics by using latent semantic alignment as a precondition for parametric knowledge transfer, thereby mitigating neural incompatibility.

\section{Motivational Analysis}
\label{sec:preliminary}

\subsection{Preliminary: Language Model Semantics}

\begin{figure}[!htb]
    \centering
    \includegraphics[width=1.0\linewidth]{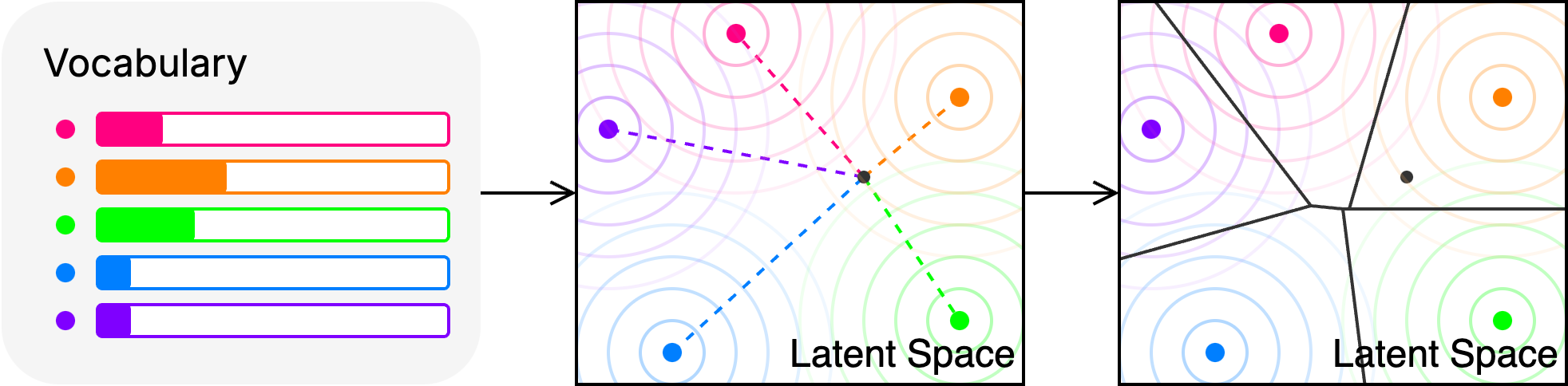}
    \caption{Semantic association of vocabulary and latent space.
    For each color label on the vocabulary (left), there is a color semantic basis in the latent space (middle). The semantics of the dark dot (indicating an arbitrary representation) in the latent space can be quantified as its cosine similarities to semantic bases. The semantics can be computed as probabilities on the vocabulary. When focusing on the nearest semantic basis for a given latent representation, a latent space can be quantified as discrete semantic regions (right).}
    \label{fig:semantics}
\end{figure}

For the recognizable semantic meanings of a given LM, \emph{vocabulary-defined semantics} proposed defining a set of special representations in the latent space to associate vocabulary labels. It quantifies semantic structure in the LM latent space by leveraging local isotropy~\citep{Cai2021IsotropyIT}, and it has been used for parameter optimizations such as efficient logits computation~\citep{gu2024vocabulary}.
For each label in the LM vocabulary, there is an associated latent representation, termed a ``semantic basis'', and they share the same semantic meaning, as shown in \cref{fig:semantics}.

For a given LM-head matrix, we conduct matrix multiplication to obtain semantic bases in the latent space.
Since the computation direction is from logits to representations, instead of using the LM-head matrix $\mathbb{W}$, we use its pseudoinverse $\mathbb{W}^+$.
If there are $v$ labels in the vocabulary, there will be $v$ unique semantic bases representing all semantic meanings.
At the output side of LM, each onehot embedding $\vec{e}$ is multiplied by the pseudoinverse matrix $\mathbb{W}^+$ to obtain the corresponding representation $\vec{s}$. That is, $\vec{s}=\vec{e}\cdot\mathbb{W}^+$.
The computation is equivalent to solving the least-squares problem of a system of linear equations. Equivalently, the LM-head and its pseudoinverse form a readout/reconstruction pair: the LM-head maps a hidden state to semantic coordinates on the vocabulary, while the pseudoinverse maps those coordinates back to a latent representation in the least-squares sense.
The time cost of computing semantic bases is rather low. For language models like LLaMA 2 (7B, 13B, and even 70B), which have 32k labels in the vocabulary, it takes around 10 seconds on A100. Moreover, this is an one-time computation with persistent value.

\subsection{Empirical Finding: Vector Nature of Latent Semantics}

Centered on each semantic basis, there forms a ``semantic field''. The concept of semantic field is analogous to the \textit{field} term in physics (such as electric field), with the semantic basis acting as the pole of the field. The semantics of an arbitrary latent representation can be quantified as the overlapping impact of multiple semantic fields, and further computed as probabilities~\citep{gu2024vocabulary}.
The process is ``composition of semantics'', where multiple \emph{semantic components} become a \emph{resultant vector} via vector addition.
We propose a hypothesis that the overlapping effects of semantic fields support a corresponding reversed operation, namely ``resolution of semantics''. That is, a single \emph{resultant vector} in latent space may be resolved into multiple \emph{component vectors} along the directions of semantic bases.

In detail, for a given latent representation $\vec{r}$, its semantic meaning can be projected to different semantic bases to obtain corresponding semantic components $\vec{c_{i}}=\mathtt{proj}(\vec{r}, \vec{s_{i}})$ (analogy to ``component force'' in a force field). By accumulating the decomposed semantics, we get a ``resultant semantics'' $\sum\limits_{i=1}^{n} \vec{c_{i}}$ (analogy to ``resultant force'' in a force field).
The equation $\vec{r} \parallel \sum\limits_{i=1}^{n} \vec{c_{i}}$ stands approximately true. In contrast, when taking a random collection of vectors as semantic bases and obtaining $\vec{c_{i}^\prime}=\mathtt{proj}(\vec{r}, \vec{s_{i}^\prime})$, the equation $\vec{r} \perp \sum\limits_{i=1}^{n} \vec{c_{i}^\prime}$ stays true. It is consistent with the property of high-dimensional latent space that arbitrary vectors tend to be orthogonal to each other.

\begin{figure}[!htb]
    \centering
    \includegraphics[width=1.0\linewidth]{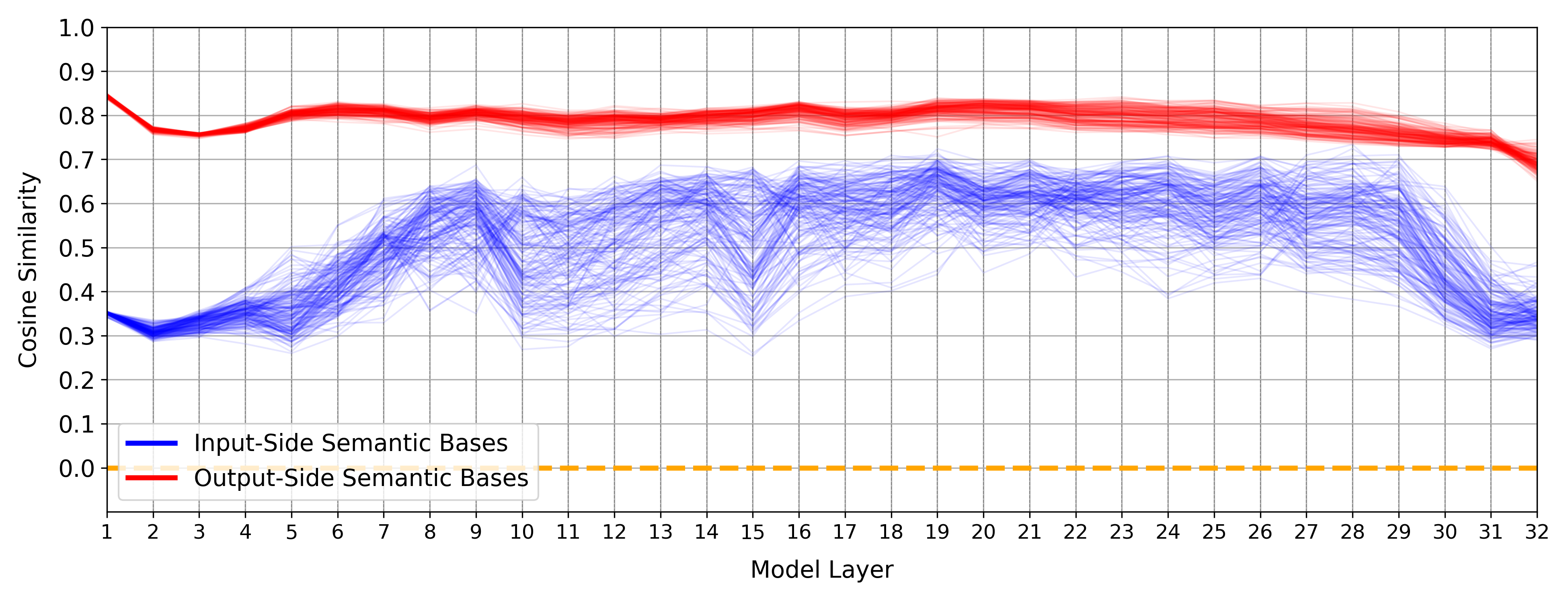}
    \caption{Validation of Vector Nature of Latent Semantics (on HumanEval with Llama2-7B).}
    \label{fig:empirical_validation}
\end{figure}

We conduct empirical experiments to validate the hypothesis. For a given dataset and LM, we first compute the outputs of each layer, and then decompose each layer output into semantic components and eventually recompose it back as a layer output. The question tested in this experiment is simple: does decomposition followed by recomposition recover a vector that keeps nearly the same semantic direction as the original hidden state? If the old layer outputs and the new layer outputs share almost the same direction in the latent space, namely if their cosine similarity is high, the hypothesis is supported.
We run Llama2-7B on HumanEval, to study whether the hypothesis holds with output-side semantic bases, using input-side semantic bases for comparison.
As shown in \cref{fig:empirical_validation}, the hypothesis holds when using output-side semantic bases because of the high cosine similarities across layers. In contrast, input-side semantic bases yield cosine similarities close to zero, consistent with the high-dimensional tendency of arbitrary vectors to be nearly orthogonal.

\section{Approach}
\label{sec:approach}

\begin{figure*}[!ht]
    \centering
    \includegraphics[width=\linewidth]{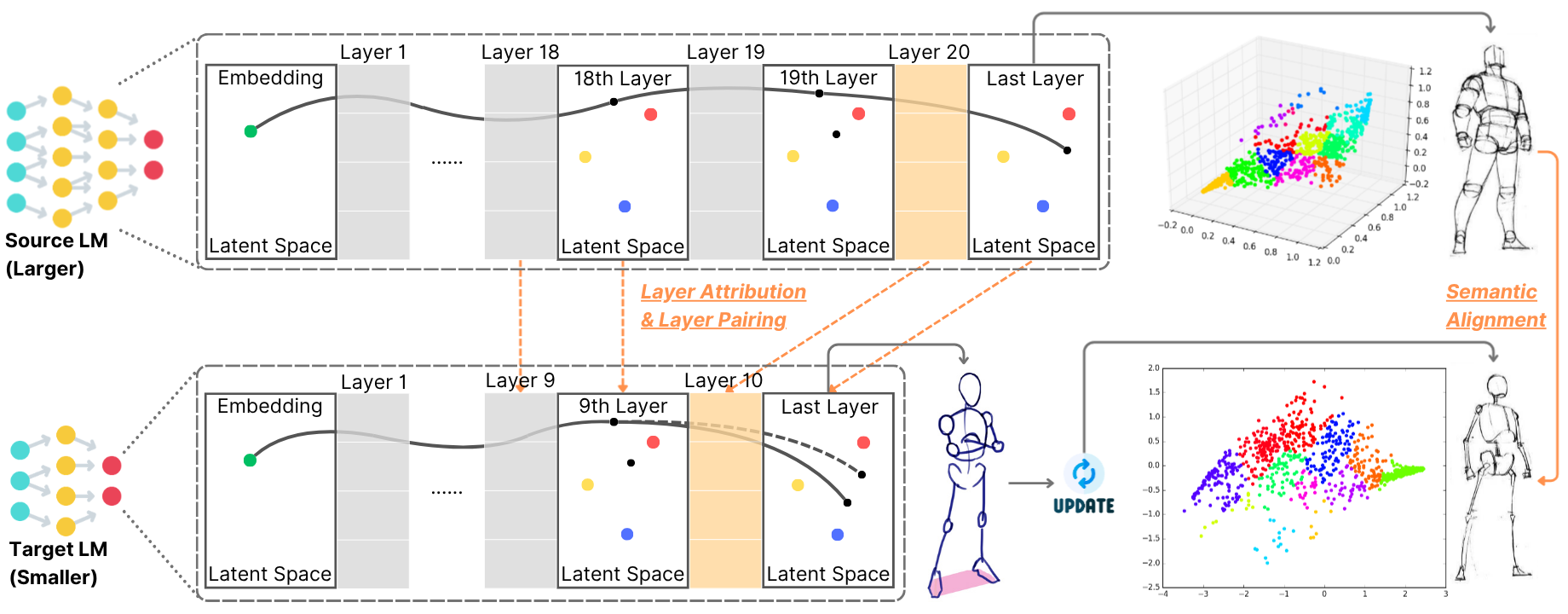}
    \caption{Illustration of our cross-scale knowledge transfer approach. Assume a 20-layer source LM and a 10-layer target LM, then:
    First, the source-side extract stage selects critical source layers, and the selected source layers are paired one-to-one with all target layers from shallow to deep. A highlighted pair is shown in orange only for illustration;
    Second, represented by the dots in 3D and 2D spaces, layer outputs from the source model are decomposed in the larger-dimensional source latent space and recomposed in the smaller-dimensional target latent space as target-space supervision. This process reduces dimension while preserving semantics, represented by the gray figures that retain the same coarse gesture while reducing details;
    Third, the current frontier target layer is optimized so that its residual relation geometry follows the supervisory signal, while the other target layers remain frozen. This is analogous to adjusting the blue figures to express the same coarse gesture as the gray figures.
    After cross-scale knowledge transfer, target layer outputs move toward the supervisory signal, represented by the dashed curve. At the output side, the dark point denotes the hidden representation whose proximity to vocabulary semantic bases determines the predicted label. The source representation is closest to the red semantic basis, and the transferred target representation is steered from the blue basis toward the same red basis.
    }
    \label{fig:illustration}
\end{figure*}

Our approach uses LM semantics to align the latent space between paired layers of the source and target models. The transferred knowledge under semantic alignment is represented by layer activations, which provide the supervisory signal for parameter optimization.
We name our approach \ul{Sem}antic \ul{Align}ment, shortened as \textsc{SemAlign}.
The illustration of our approach is in \cref{fig:illustration}, and the main procedure is as follows:
(1) we locate critical layers in the source LM by attribution and select exactly one source layer for each target layer; these selected source layers are then paired with all target layers from shallow to deep;
(2) we decompose the semantics of source layer outputs in the source latent space and recompose them as target-space supervisory states;
(3) in the target model, we optimize paired target layers stage by stage. At each stage, the trainable layer is supervised at the residual level through centered token-token relation matching, while an output KL term preserves source predictive behavior. In this way, the target model gradually reproduces the source behavior through aligned activations rather than transferred source parameters.

\subsection{Layer Attribution and Pairing}

\paragraph{Layer Attribution}
We identify candidate source layers using \emph{Layer Gradient~$\times$~Activation} at the layer level: for each layer, we multiply the gradient of the supervised objective by the layer's activations and aggregate over tokens and channels to obtain a scalar importance score per layer. This choice is simple, stable, and available in standard attribution toolkits \citep{kokhlikyan2020captumunifiedgenericmodel}.

\paragraph{Layer Pairing}
Let the source have $L_{\mathrm{src}}$ layers and the target have $L_{\mathrm{tgt}}$ layers. The extract stage scores source layers on the extract split and selects exactly $L_{\mathrm{tgt}}$ source layers from the source LM only. Let their sorted indices be
$\mathcal{S}_{\mathrm{src}} = \{ s_1 < s_2 < \cdots < s_{L_{\mathrm{tgt}}} \}$.
We then pair them directly with all target layers from shallow to deep: $(s_k,\,k)$, $\qquad k\in\{1,\dots,L_{\mathrm{tgt}}\}$.
This rule removes the need to locate critical target layers and gives a simple one-to-one correspondence for the subsequent semantic-alignment stage. In other words, attribution decides \emph{which} source layers to transfer, and ordering decides \emph{where} they supervise in the target.

\subsection{Latent Semantic Alignment}

We construct a training-free, semantics-preserving mapping from a source layer's latent space to a target layer's latent space. In this paper, the source and target come from the same LM family at different scales, so their vocabularies and output-side semantic bases are index-aligned (Section~\ref{sec:preliminary}). The idea is to \emph{decompose} a source vector into semantic components and \emph{recompose} those components in the target basis with the same coefficients.

Let $\mathbf{h}^{\mathrm{src}}\in\mathbb{R}^{D_{\mathrm{src}}}$ be a source-layer output at the supervision interface. Let the source and target semantic bases be
\begin{equation}
\label{eq:bases}
\begin{aligned}
S_{\mathrm{src}}
&= [\mathbf{s}^{\mathrm{src}}_1,\dots,\mathbf{s}^{\mathrm{src}}_m]\in\mathbb{R}^{D_{\mathrm{src}}\times m},\\
S_{\mathrm{tgt}}
&= [\mathbf{s}^{\mathrm{tgt}}_1,\dots,\mathbf{s}^{\mathrm{tgt}}_m]\in\mathbb{R}^{D_{\mathrm{tgt}}\times m},
\end{aligned}
\end{equation}
where column $i$ in $S_{\mathrm{src}}$ and $S_{\mathrm{tgt}}$ refers to the same semantic atom. The semantic bases themselves need not be unit-normalized; in the practical implementation we keep the raw output-side bases induced by the LM head and its pseudoinverse.

\paragraph{Semantic decomposition and recomposition}
For conceptual exposition, we write the semantic coefficients of the source representation in cosine style:
\begin{equation}
\label{eq:source-cos}
a_i = \cosim(\mathbf{h}^{\mathrm{src}},\mathbf{s}^{\mathrm{src}}_i),
\qquad i\in\{1,\dots,m\}.
\end{equation}
Collecting them gives $\mathbf{a}=[a_1,\dots,a_m]^\top\in\mathbb{R}^m$. We then recompose these coefficients in the target space:
\begin{equation}
\label{eq:target-recompose}
\tilde{\mathbf{h}}^{\mathrm{tgt}}
= S_{\mathrm{tgt}}\,\mathbf{a}
\in\mathbb{R}^{D_{\mathrm{tgt}}}.
\end{equation}
The vector $\tilde{\mathbf{h}}^{\mathrm{tgt}}$ serves as the target-space supervisory state for the paired layer. Because the same coefficient vector $\mathbf{a}$ is used on index-aligned semantic bases, the semantic content is preserved while only the ambient basis changes. In realization, this decomposition and recomposition view is realized with the raw output-side semantic bases, rather than by explicitly normalizing the basis vectors. For batched token representations, the mapping is applied token-wise to valid supervision positions.

\subsection{Residual Relation Loss}
\label{sec:rep-align}

In the second stage, we adapt the target with a shallow-to-deep frontier schedule. At frontier stage $k$, only the current target layer $k$ is trainable, while all other target layers are frozen. The layer loss supervises what the active target layer contributes, rather than the whole hidden state at that depth. This residual-level supervision separates the local transfer signal from the accumulated target context and avoids simultaneous interference from many layer losses. All quantities in this subsection are computed only on valid, non-padding tokens used for next-token supervision.

Let $\mathbf{x}^{\mathrm{tgt}}_k\in\mathbb{R}^{B\times T\times D_{\mathrm{tgt}}}$ be the input to target layer $k$, and let $\mathbf{y}^{\mathrm{tgt}}_k$ be the corresponding output after the active layer. Let $\tilde{\mathbf{y}}^{\mathrm{src}\rightarrow\mathrm{tgt}}_k$ denote the target-space supervisory state obtained by applying semantic decomposition and recomposition to the output of the paired source layer $s_k$. We define the target residual and the supervisory residual as
\begin{equation}
\label{eq:residual-fields}
\begin{aligned}
\Delta^{\mathrm{tgt}}_k
&= \mathbf{y}^{\mathrm{tgt}}_k - \mathbf{x}^{\mathrm{tgt}}_k,\\
\Delta^{\star}_k
&= \tilde{\mathbf{y}}^{\mathrm{src}\rightarrow\mathrm{tgt}}_k - \mathbf{x}^{\mathrm{tgt}}_k.
\end{aligned}
\end{equation}
Thus the active layer is trained to contribute the residual displacement that moves the target input toward the aligned source supervision in the target latent space. The supervisory residual is defined relative to the current target-layer input, which keeps the objective local to the active target layer. The supervision is not an absolute-state match: the loss is applied to the geometry of token-wise residual fields.

For each sequence, we remove common-mode drift across tokens before relation matching. With a valid-token mask $m_i\in\{0,1\}$, let
\begin{equation}
\label{eq:center-residuals}
\begin{aligned}
\hat{\Delta}^{\mathrm{tgt}}_{k,i}
&= \Delta^{\mathrm{tgt}}_{k,i}
- \frac{\sum_j m_j\Delta^{\mathrm{tgt}}_{k,j}}{\sum_j m_j},\\
\hat{\Delta}^{\star}_{k,i}
&= \Delta^{\star}_{k,i}
- \frac{\sum_j m_j\Delta^{\star}_{k,j}}{\sum_j m_j}.
\end{aligned}
\end{equation}
We then normalize each centered residual vector and compare signed token-token relations:
\begin{equation}
\label{eq:relation-kernels}
\begin{aligned}
K^{(k)}_{ij}
&= \left\langle
\frac{\hat{\Delta}^{\mathrm{tgt}}_{k,i}}{\|\hat{\Delta}^{\mathrm{tgt}}_{k,i}\|_2},
\frac{\hat{\Delta}^{\mathrm{tgt}}_{k,j}}{\|\hat{\Delta}^{\mathrm{tgt}}_{k,j}\|_2}
\right\rangle,\\
K^{\star(k)}_{ij}
&= \left\langle
\frac{\hat{\Delta}^{\star}_{k,i}}{\|\hat{\Delta}^{\star}_{k,i}\|_2},
\frac{\hat{\Delta}^{\star}_{k,j}}{\|\hat{\Delta}^{\star}_{k,j}\|_2}
\right\rangle.
\end{aligned}
\end{equation}
The sign of $K^{\star(k)}_{ij}$ is kept in the supervisory relation, while its magnitude provides confidence for the pair. The centered residual relation loss is
\begin{equation}
\label{eq:layer-relation}
\begin{aligned}
\mathcal{L}_{\mathrm{rel}}^{(k)}
&=
\frac{
\sum_{i\neq j} w^{(k)}_{ij}\bigl(K^{(k)}_{ij}-K^{\star(k)}_{ij}\bigr)^2
}{
\sum_{i\neq j} w^{(k)}_{ij}
},\\
w^{(k)}_{ij}
&=m_i m_j\left|K^{\star(k)}_{ij}\right|.
\end{aligned}
\end{equation}
This objective focuses on relative token structure rather than sequence-level bias. Pair weights use relation strength as confidence and the valid-token mask excludes padding, but the signed relation remains the supervision target.

At the output layer, let $\mathbf{p} = \softmax(\mathbf{z})$, we optimize
\begin{equation}
\label{eq:out-kl}
\mathcal{L}_{\mathrm{out}}
= \operatorname{Avg}\!\left[ \KL\!\left(\mathbf{p}^{\mathrm{src}}\,\|\,\mathbf{p}^{\mathrm{tgt}}\right) \right].
\end{equation}
At frontier stage $k$, the total objective is
\begin{equation}
\label{eq:total-loss}
\mathcal{L}^{(k)}
=\mathcal{L}_{\mathrm{rel}}^{(k)} + \mathcal{L}_{\mathrm{out}}.
\end{equation}
Here $\operatorname{Avg}[\cdot]$ denotes averaging over valid supervised next-token positions. After the current frontier stage is optimized, we freeze layer $k$ and move to layer $k{+}1$, so the semantic-alignment stage proceeds from shallow to deep.

The layer term is a representation-alignment objective on residual geometry. It differs from conventional intermediate-layer distillation, which typically matches source features or source logits directly in the source space. Here we first transport the supervision into the target semantic space and then compare the relational structure of the residual contribution made by the active target layer. The output KL term has the usual distillation interpretation and anchors the whole model to source-level predictions, while the local relation loss shapes internal transfer.

\section{Experiments}
\label{sec:experiments}

In the experiments, we mainly study how our approach performs in parametric knowledge transfer compared with PKT baselines. We also conduct analysis based on the latent representation similarities between source and target models before and after latent semantic alignment.

\subsection{Setup}

\paragraph{Datasets}
The evaluations are with four benchmarks that cover representative downstream capabilities:
MMLU evaluates broad professional and academic knowledge \citep{hendrycks2020measuring};
GSM8K evaluates grade-school mathematical reasoning \citep{cobbe2021training};
and HumanEval and MBPP evaluate functional code generation \citep{chen2021evaluating,austin2021program}.
For each benchmark, the data used for transfer are separated from benchmark evaluation.
Specifically, we use an \emph{extract split} only for source-layer attribution and an \emph{align split} only for semantic alignment and knowledge transfer.
The reported results are obtained using the evaluation protocol of each benchmark.

\paragraph{Models}
We conduct experiments with Llama2 \citep{touvron2023llama} models, mainly chat versions instead of base versions for better instruction-following ability.
Besides, we employ LM variants to study transfer from further-finetuned source models to the same target model,
CodeLlama-13B-Python~\citep{rozière2024codellamaopenfoundation} and WizardCoder-13B-Python~\citep{luo2023wizardcoder}.
They are finetuned on Llama2-13B with massive code data for enhanced coding performance.

\paragraph{Metrics}
For MMLU and GSM8K, we calculate \emph{accuracy} in the zero-shot setting;
and for HumanEval and MBPP, we calculate \emph{pass@1}.
Larger scores mean better performance.

\paragraph{Baselines}
We compare against the two representative PKT baselines, \textsc{Seeking} and \textsc{LaTen}. \textsc{Seeking} follows a PostPKT pipeline with extraction, injection, and post-alignment finetuning. \textsc{LaTen} follows a PrePKT pipeline with localization, pre-alignment, and injection. In contrast, SemAlign uses a two-stage workflow: source-side extract followed by a merged semantic-alignment stage with shallow-to-deep frontier optimization.

\subsection{Results}

\paragraph{Cross-Scale Knowledge Transfer}
According to the results shown in \cref{tab:overall}, \textsc{SemAlign} improves substantially over Llama2-7B-Chat while remaining below the Llama2-13B-Chat source, and it stays closer to the source than the other transfer baselines on average. Concretely, the absolute gaps between SemAlign and 13B are $2.60$ on MMLU, $1.34$ on GSM8K, $1.41$ on HumanEval, and $0.42$ on MBPP, averaging $1.44$ points. This average gap is smaller than \textsc{Seeking} ($\approx 3.92$) and \textsc{LaTen} ($\approx 3.43$). At the task level, LaTen is numerically closest to the source on GSM8K, while \textsc{Seeking} overshoots the source on the same task ($28.23$). Overall, SemAlign shows three closer margins on four benchmarks, which indicates that it learns source behavior more faithfully than the baselines.

\begin{table}[htbp]
\centering
\resizebox{\linewidth}{!}{%
\begin{tabular}{lcccc}
\toprule
Models & MMLU & GSM8K & HumanEval & MBPP \\
\midrule
Llama2-7B-Chat & 44.20 & 16.07 & 14.05 & 17.80 \\
\rowcolor{gray!20} Llama2-13B-Chat & 52.90 & 20.55 & 18.75 & 19.20 \\
\quad Seeking & 49.60 & \textbf{28.23} & 15.44 & \textbf{20.60} \\
\quad LaTen & 44.40 & 20.47 & 14.63 & 18.20 \\
\quad \textbf{SemAlign} & \textbf{50.30} & 19.21 & \textbf{17.34} & 18.78 \\
\bottomrule
\end{tabular}
}
\caption{Results of Parametric Knowledge Transfer in Downstream Tasks.}
\label{tab:overall}
\end{table}

At the task level, SemAlign achieves the best transferred performance on MMLU ($50.30$) and HumanEval ($17.34$), surpassing both Seeking ($49.60$, $15.44$) and \textsc{LaTen} ($44.40$, $14.63$), whereas \textsc{Seeking} leads on GSM8K ($28.23$) and MBPP ($20.60$). This result pattern is also consistent with the design of the methods. In SemAlign, the target is optimized with supervisory hidden states constructed from the source model, together with the source output distribution. The objective is therefore to reproduce source behavior as faithfully as possible, rather than to perform an additional stage of task-specific finetuning. By contrast, \textsc{Seeking} includes a post-alignment finetuning stage after injection, which naturally gives it more room to overshoot the source on some tasks. For the same reason, it is difficult for a transferred target produced by SemAlign to consistently outperform the source model itself.

A clear pattern is the trade-off between stability and aggressiveness across methods. \textsc{Seeking} exceeds the 13B source on both GSM8K ($+7.68$) and MBPP ($+1.40$), suggesting stronger task-specific optimization. SemAlign remains below but close to the source, and stays within $2.60$ points of the source on every task. A second plausible factor is that the realization uses model-wise output-side semantic bases as a practical approximation to layer-wise semantics; this approximation is robust overall, but it may be less precise on tasks that require especially sharp token-level distinctions.

\paragraph{Knowledge Transfer from Finetuned Models}
Based on the results shown in \cref{tab:finetuned}, in five of six settings, \textsc{SemAlign} consistently outperforms the transfer baselines, indicating stronger parametric knowledge transfer. With Llama2-13B-Chat as source, it leads \textsc{LaTen} and \textsc{Seeking} on HumanEval, and only trails \textsc{Seeking} on MBPP. The advantage becomes clearer with code-specialized sources: from CodeLlama-13B-Python, SemAlign reaches $20.12$ ($+4.07$ over \textsc{Seeking}, $+6.10$ over \textsc{LaTen}) on HumanEval and $22.35$ ($+0.95$, $+4.55$) on MBPP; from WizardCoder-13B-Python, it attains $19.46$ ($+4.42$, $+5.44$) and $21.18$ ($+1.38$, $+2.58$) on HumanEval and MBPP, respectively. These trends show that SemAlign extracts and transfers competency from finetuned models more reliably than the baselines.

\begin{table}[htbp]
\centering
\resizebox{0.9\linewidth}{!}{%
\begin{tabular}{lcc}
\toprule
Models & HumanEval & MBPP \\
\midrule
Llama2-7B-Chat & 14.05 & 17.80 \\
\rowcolor{gray!20} Llama2-13B-Chat & 18.75 & 19.20 \\
\quad Seeking & 15.44 & \textbf{20.60} \\
\quad LaTen & 14.63 & 18.20 \\
\quad \textbf{SemAlign} & \textbf{17.34} & 18.78 \\
\rowcolor{gray!20} CodeLlama-13B-Python & 47.56 & 37.80 \\
\quad Seeking & 16.05 & 21.40 \\
\quad LaTen & 14.02 & 17.80 \\
\quad \textbf{SemAlign} & \textbf{20.12} & \textbf{22.35} \\
\rowcolor{gray!20} WizardCoder-13B-Python & 56.71 & 41.60 \\
\quad Seeking & 15.04 & 19.80 \\
\quad LaTen & 14.02 & 18.60 \\
\quad \textbf{SemAlign} & \textbf{19.46} & \textbf{21.18} \\
\bottomrule
\end{tabular}
}
\caption{Results of Parametric Knowledge Transfer from Finetuned Source Models.}
\label{tab:finetuned}
\end{table}

Across the two coding benchmarks, all methods remain far below the finetuned sources (CodeLlama-13B-Python: $47.56$ on HumanEval and $37.80$ on MBPP; WizardCoder-13B-Python: $56.71$ and $41.60$), despite often surpassing Llama2-7B-Chat and sometimes even matching or exceeding Llama2-13B-Chat. This gap suggests that the extensive, task-specific optimization embedded in code-specialized sources is difficult to reconstruct via short-horizon transfer; semantic-coordinate matching encourages conservative alignment to source representations rather than aggressive task re-optimization, limiting the attainable ceiling without longer or more targeted finetuning.

An additional observation is that \textsc{Seeking} only overshoots the source on MBPP when the source is the generalist Llama2-13B-Chat ($20.60 > 19.20$), but not when the source is code-specialized. Meanwhile, \textsc{SemAlign} shows its largest margins over baselines precisely when transferring from code-specialized sources. This pattern hints that aggressive, task-specific optimization in \textsc{Seeking} can exploit headroom left by generalist sources, whereas SemAlign's representation-faithful transfer scales better with source specialization.

\subsection{Analysis on Feature Representations}

We adopt Centered Kernel Alignment (CKA)~\citep{kornblith2019similarity} as the analysis tool to study the similarities between layer outputs from source and target models.
We run Llama2-chat models on HumanEval data.
CKA is commonly used to compute the similarities between feature representations in neural networks, which is based on Hilbert-Schmidt Independence Criterion (HSIC).

\begin{figure}[ht]
    \centering
    \begin{subfigure}[b]{\linewidth}
        \centering
        \includegraphics[width=\linewidth]{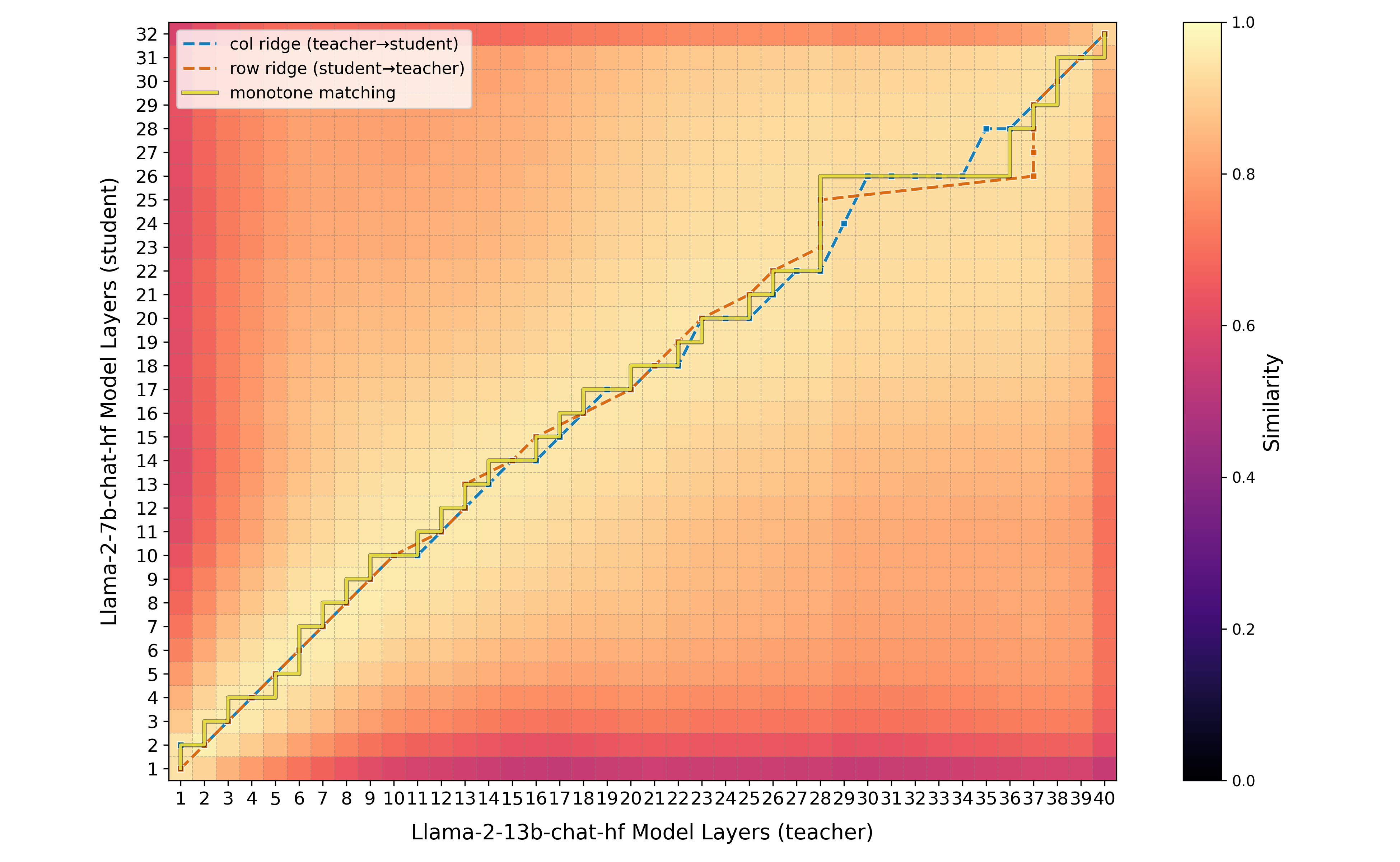}
        \caption{source-target without alignment}
    \end{subfigure}
    \hfill
    \begin{subfigure}[b]{\linewidth}
        \centering
        \includegraphics[width=\linewidth]{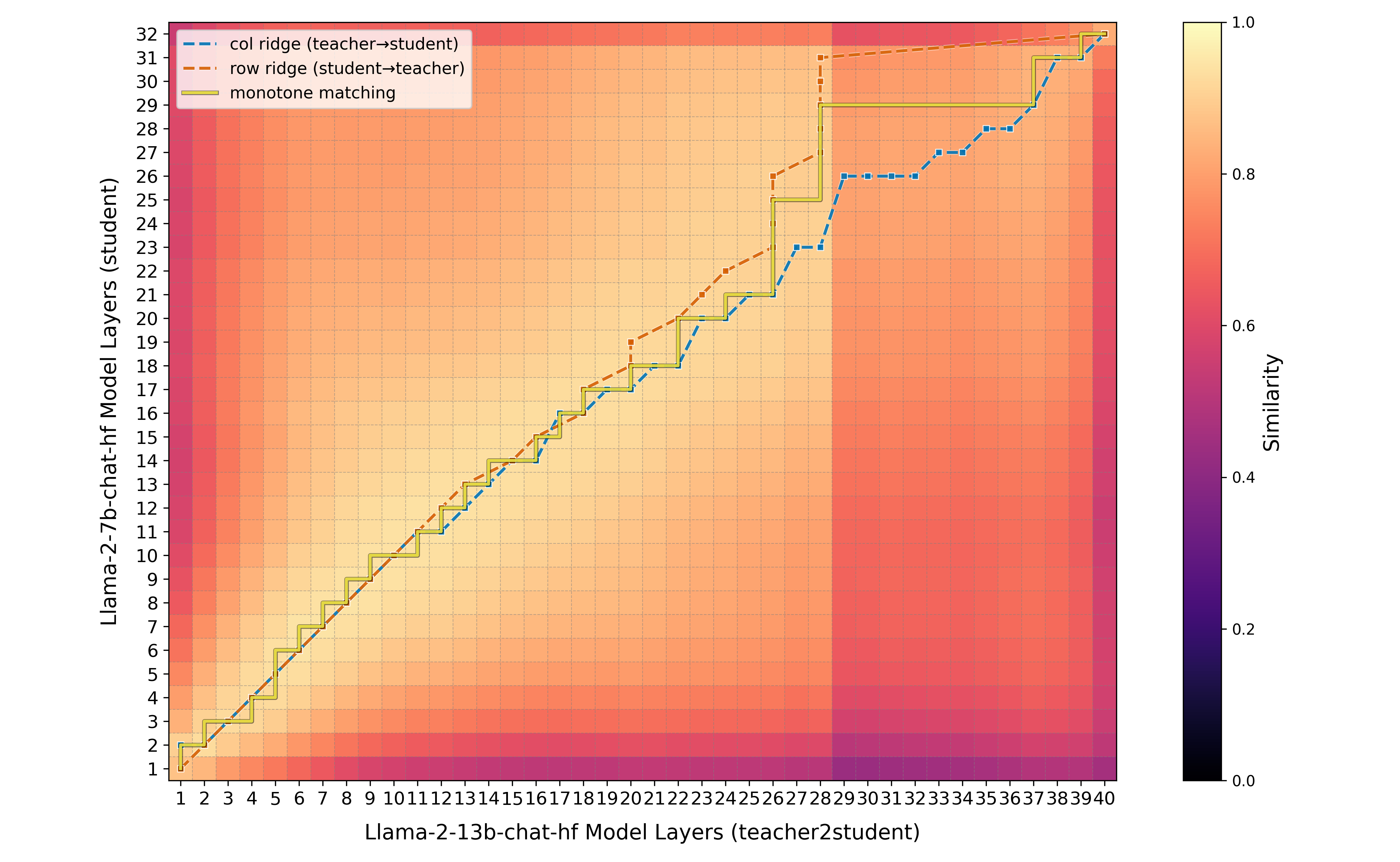}
        \caption{source-target with alignment}
    \end{subfigure}
    \caption{Comparison of Layer-wise Representation Similarities between Cross-Scale LMs.}
    \label{fig:representation_similarity}
\end{figure}

As shown in~\cref{fig:representation_similarity}, there are high similarities between the layer outputs from source and target models, especially along the main diagonal. This indicates that there is no strong behavioral incompatibility when layer outputs are used as the medium of parametric knowledge transfer, instead of directly using layer parameters. The highest similarities are almost layer by layer, from shallow to deep.
Meanwhile, the cases before and after latent semantic alignment share a very similar pattern of similarities. This suggests that latent semantic alignment is a safe way to utilize the cross-scale similarities between layer outputs.

\section{Conclusion}
We studied parametric knowledge transfer across differently scaled LMs from a \emph{semantics-first} perspective. Rather than moving raw parameters as in prior paradigms, we use layer outputs as the medium of transfer and identify \emph{latent semantic alignment} as the prerequisite for stable cross-scale transfer.
Building on this view, \textsc{SemAlign} first extracts a source-only set of critical layers and then performs a semantic-alignment stage that pairs them with all target layers from shallow to deep, decomposes and recomposes their latent semantics, and steers the target stage by stage with target-space supervisory states. The semantic-alignment stage follows a shallow-to-deep frontier schedule in which only the current target layer is trainable. Its local objective matches centered residual relation geometry, while output KL remains active throughout to preserve predictive behavior. This design avoids neural incompatibility, simplifies the procedure, and makes transfer efficient in both compute and data.
Empirically, SemAlign improves targets over task-matched baselines and over parameter-space transfer methods on four benchmarks with Llama~2 families. The results support our central claim: treating activations as the carrier of knowledge and aligning semantics first provides a robust path to cross-scale PKT.

\section*{Limitations}

Our evaluation is limited in the scope. The experiments are on four benchmarks and mainly study transfer within the Llama-2 family across scales, with a few further-finetuned variants. It is still not clear how well SemAlign works for more diverse settings, such as other model families with different tokenizers or architectures, longer-context use, multilingual tasks, or safety-critical behavior.

SemAlign further assumes that the source and target share a usable semantic correspondence, so that we can decompose a representation in the source and recompose it in the target. When this correspondence is weak, such as under vocabulary mismatch or semantic drift, the alignment signal can become noisy and transfer may degrade. Extending the method to cross-family transfer will likely require additional transformations on semantic bases and explicit treatment of vocabulary mismatch, which we leave to future work. The implementation also uses model-wise output-side semantic bases as a practical approximation to layer-wise semantics, and this approximation may be less precise on tasks that require especially sharp token-level distinctions.

More broadly, our method operates at the layer level and uses lightweight updates, which keep the two-stage procedure simple and stable but can cap the best achievable performance and do not fully prevent interference with other skills. The frontier schedule reduces cross-layer interference during training, but it also makes the transfer procedure more conservative than full joint optimization.


\bibliography{custom}

\appendix

\section{Technical Details}
\label{appendix:details}

\subsection{Stats of Language Models}

The stats of language models in our experiments are shown in \cref{tab:model_stats}.

\begin{table}[!ht]
    \centering
    \resizebox{\linewidth}{!}{%
    \sisetup{table-format=2.2}
\rowcolors{2}{}{gray!10}
\begin{tabular}{
    l cc c c
}

\hiderowcolors
\toprule

& \multicolumn{2}{c}{Llama2} & {CodeLlama} & {WizardCoder} \\
\cmidrule(lr){2-3} \cmidrule(lr){4-4} \cmidrule(lr){5-5}
& 7B & 13B & 13B & 13B \\
\midrule


Head Num. & 32 & 40 & 40 & 40 \\

Layer Num. & 32 & 40 & 40 & 40 \\

Dimension & 4,096 & 5,120 & 5,120 & 5,120 \\

Vocabulary & 32,000 & 32,000 & 32,000 & 32,001 \\

\bottomrule

\end{tabular}

    }
    \caption{Stats of Llama2 Language Models.}
    \label{tab:model_stats}
\end{table}

\subsection{Implementation Details}

We follow the experimental protocol of \textsc{LaTen} for a fair comparison, but the SemAlign workflow has two stages only.
The \emph{extract} stage uses the extract split to attribute source layers on the source LM only, and it selects exactly $L_{\mathrm{tgt}}$ source layers for a target with $L_{\mathrm{tgt}}$ layers.
The \emph{semantic-alignment} stage then pairs the selected source layers with all target layers from shallow to deep and optimizes the target on the align split through a frontier schedule.
At frontier stage $k$, only target layer $k$ is trainable and all other target layers are frozen.

For each active layer, we construct the target-space supervisory state in Section~\ref{sec:approach}. We compute the target residual contribution $\Delta^{\mathrm{tgt}}_k$ from the active target layer and the supervisory residual $\Delta^\star_k$ from the aligned source supervision relative to the same target-layer input. The two residual fields are centered across valid tokens, normalized token-wise, and compared through the weighted signed relation loss in Section~\ref{sec:rep-align}. At every frontier stage, we also keep the KL loss between the source and target output distributions, and the two terms are combined as in Eq.~\ref{eq:total-loss}.
We update only LoRA parameters inserted into FFN (\texttt{up\_proj}, \texttt{down\_proj}) and MHSA (\texttt{v\_proj}, \texttt{o\_proj}) modules, with rank $r{=}16$, and only for the active frontier layer. Both terms are computed on the standard shifted next-token supervision positions and averaged over valid tokens. For the optimization schedule of the semantic-alignment stage, we mainly adopt the hyperparameters of \textsc{Seeking}'s finetuning stage. The experiments are conducted via a single run, with the global random seed $42$.
Our implementation uses deep learning framework \textsc{PyTorch}~\citep{Paszke2019PyTorchAI}, \textsc{Transformers}~\citep{Wolf2020TransformersSN}, and vLLM-backed \textsc{LM-Eval}~\citep{kwon2023efficient}.

\begin{table}[htbp]
\centering
\resizebox{0.88\linewidth}{!}{%
\begin{tabular}{lccc}
\toprule
& Epochs & Batch Size & Learning Rate \\
\midrule
MMLU & 5 & 16 & 3e-4 \\
GSM8K & 5 & 16 & 3e-4 \\
HumanEval & 3 & 16 & 3e-5 \\
MBPP & 5 & 16 & 3e-4 \\
\bottomrule
\end{tabular}
}
\caption{Optimization Details in Semantic Alignment.}
\label{tab:details_align}
\end{table}

\subsection{AI Use Disclosure}

We use a business LLM (GPT-5~\footnote{https://openai.com/gpt-5/}) to merely aid or polish our paper writing.

\section{Discussion}
\label{appendix:discussion}

\subsection{Medium for Parametric Knowledge Transfer}
Compared with prior work such as Seeking and LaTen, which use model weights as the medium for knowledge transfer, \textsc{SemAlign} uses layer outputs as the transfer medium.
This choice improves efficacy and requires only limited additional computation for the alignment map.
It is also preferable from an information-transfer perspective.

Layer outputs require less transfer bandwidth than layer parameters or full vocabulary distributions.
For language models, the dimension of a layer output is much smaller than the number of parameters in a layer and also the vocabulary-sized probability vector.
A smaller transfer medium reduces the amount of information that must be transported, making the procedure more efficient and less exposed to cross-space mismatch.

Layer outputs are also safer than direct parameter manipulation from the perspective of knowledge-neuron association.
Prior work shows that LM knowledge and neurons follow many-to-many dynamic associations~\citep{allenphysics}.
If transfer is conducted through layer parameters, certain target parameters will be changed according to certain source parameters, even though both parameter sets also participate in many other computations outside the transfer data.
By contrast, SemAlign uses activations produced on the transfer data as the supervision target and updates only lightweight adapters in the active frontier layer.

\subsection{Residual Geometry and Frontier Optimization}

The semantic-alignment stage transfers target-space residual geometry rather than raw parameters or absolute hidden states. Using the notation of Section~\ref{sec:rep-align}, the layer loss compares the target residual and the aligned supervisory residual:
\[
\begin{aligned}
\Delta^{\mathrm{tgt}}_k
&=\mathbf{y}^{\mathrm{tgt}}_k-\mathbf{x}^{\mathrm{tgt}}_k,\\
\Delta^\star_k
&=\tilde{\mathbf{y}}^{\mathrm{src}\rightarrow\mathrm{tgt}}_k-\mathbf{x}^{\mathrm{tgt}}_k.
\end{aligned}
\]
This formulation asks the active target layer to contribute the displacement that moves its own input toward the aligned source supervision. It therefore avoids forcing the full hidden state to match an external representation directly.

Centered relation matching further removes sequence-level residual drift before comparing token structure. After centering and normalizing token residuals, SemAlign forms token-token relation matrices $K$ and $K^\star$ and minimizes their squared discrepancy. The signed supervisory relation is preserved, while the relation magnitude is used as confidence through the pair weight. In this form, the layer objective focuses on whether the active layer induces the same relative token geometry as the aligned supervision.

The frontier schedule complements this residual objective. Since only one target layer is trainable at a time, the local residual loss is not damaged by simultaneous updates to many layers. The output KL term remains active at every stage and anchors the whole model to source-level predictive behavior. Together, the residual relation loss and the output KL term provide a stable division of labor: the local objective shapes internal semantic transfer, and the output objective protects behavior at the prediction level.

\section{Details of Parametric Knowledge Transfer Baselines}

This appendix summarizes the two PKT baselines that are most relevant to our comparison. Both methods transfer parameter-side information from a larger source model to a smaller target model, but they differ in where alignment happens. In the terminology of \citep{tan2025neural}, \textsc{Seeking} is a representative \emph{PostPKT} method, while \textsc{LaTen} is a representative \emph{PrePKT} method. The most relevant cross-scale comparisons in these works are reported on the same four benchmarks used in our experimental settings, namely MMLU, GSM8K, HumanEval, and MBPP.

\subsection{Illustrate \textsc{Seeking}}

\textsc{Seeking} transfers knowledge through a three-step PostPKT pipeline: \emph{extract}, \emph{inject}, and \emph{post-align} \citep{zhong2024seeking}. For a task $\mathcal{T}$ and a small seed set decoded by the source model, it scores each source parameter $\theta_i$ by sensitivity,
\[
\begin{aligned}
S^\mathcal{T}_{i,j}
&=\Big|\theta_i^\top \nabla_{\theta_i}\mathcal{L}\bigl(x^\mathcal{T}_j,y^\mathcal{T}_j\mid\Theta\bigr)\Big|,\\
S^\mathcal{T}_i
&=\sum_{j=1}^{k}S^\mathcal{T}_{i,j},
\end{aligned}
\]
and then aggregates these scores within each layer. The selected source weights are reduced to target-compatible submatrices,
\[
W^{l}_{\mathrm{extract}} = \arg\max_{W'\subseteq W^l}\; \sum_{\theta_i\in W'} S_i,
\]
then factorized with SVD to initialize a rank-$r$ LoRA pair for the target. After injection, the target is further optimized in post-alignment finetuning.

For our comparison, the important point is that \textsc{Seeking} does not stop at transfer. Its post-alignment stage can provide stronger task-specific optimization than a source-faithful transfer objective, which helps explain why it sometimes overshoots the source model on individual tasks. In the broader evaluation of \citep{zhong2024seeking}, the method is tested on several task families.

\subsection{Illustrate \textsc{LaTen}}

\textsc{LaTen} is proposed in the neural incompatibility study of \citep{tan2025neural} as a PrePKT method. It follows three steps: \emph{locate}, \emph{align}, and \emph{inject}. Given a source model $M_\ell$ with parameters $\Theta_\ell$ and a target model $M_s$ with parameters $\Theta_s$, it first localizes informative neurons and extracts source deltas, then maps them into the target parameter space with a lightweight hypernetwork. The core pipeline can be written as three short updates:
\[
\Delta\Theta_\ell \leftarrow \mathrm{Locate}(M_\ell;\mathcal{D}_{\mathrm{extract}})
\]
\[
\widehat{\Delta\Theta}_s \leftarrow g_\phi(\Delta\Theta_\ell)
\]
\[
\Theta_s^\star \leftarrow \mathrm{Inject}(\Theta_s,\widehat{\Delta\Theta}_s)
\]
The locate stage uses static neuron-level attribution over FFN and MHSA modules. The align stage trains a small two-layer MLP hypernetwork on a tiny alignment set while the base weights remain frozen. After that, the aligned deltas are injected once, with no following finetuning stage.

The key conclusion of \citep{tan2025neural} is that both PostPKT and PrePKT remain unstable across scales because of neural incompatibility. Their experiments on MMLU, GSM8K, HumanEval, and MBPP show that pre-alignment reduces some transfer cost, but does not fully resolve the mismatch between source and target parameter spaces. This is the most direct motivation for our semantics-first alternative: we also treat alignment as the prerequisite, but we align in latent semantic space rather than in parameter space.

\end{document}